\documentclass[sigconf]{acmart}

\setcopyright{none}
\renewcommand\footnotetextcopyrightpermission[1]{}
\usepackage{float}
\usepackage{array}
\usepackage{booktabs}
\usepackage{graphicx}
\usepackage{amsmath}
\usepackage{amsfonts}
\usepackage{amssymb}
\usepackage{dsfont}
\usepackage{amsthm}
\usepackage[ruled,vlined,linesnumbered]{algorithm2e}
\usepackage{enumerate}
\usepackage{subfigure}
\usepackage{multirow}
\usepackage{gensymb}

\newcommand{\figref}[1]{Figure~\ref{#1}} 
\newtheorem{problem}{Problem}
\newcommand{\reminder}[1]{\textbf{\color{red}[** #1 **]}}  
\newcommand{\hide}[1]{} 
\newcommand{\vpara}[1]{\vspace{0.1in}\noindent\textbf{#1 }}
\newcommand{\para}[1]{\vspace{0.0in}\noindent\textbf{#1 }}
\newcommand{\beq}[1]{\vspace{-0.00in}\begin{equation}#1\end{equation}\vspace{-0.00in}}

\newcommand{\besp}[1]{\begin{split}#1\end{split}}

\begin{document}

\title{
	A Probabilistic Framework for Location Inference from Social Media
}

\author{Yujie Qian}
\affiliation{
  \institution{Massachusetts Institute of Technology}
}
\email{yujieq@csail.mit.edu}

\author{Jie Tang}
\affiliation{
  \institution{Tsinghua University}
}
\email{jietang@tsinghua.edu.cn}

\author{Zhilin Yang}
\affiliation{
  \institution{Carnegie Mellon University}
}
\email{zhiliny@cs.cmu.edu}

\author{Binxuan Huang}
\affiliation{
  \institution{Carnegie Mellon University}
}
\email{binxuanh@andrew.cmu.edu}

\author{Wei Wei}
\affiliation{
  \institution{Carnegie Mellon University}
}
\email{weiwei@cs.cmu.edu}

\author{Kathleen M. Carley}
\affiliation{
  \institution{Carnegie Mellon University}
}
\email{kathleen.carley@cs.cmu.edu}

\begin{abstract}
	We study the extent to which we can infer users' geographical locations from social media.
	Location inference from social media can benefit many applications, such as disaster management, targeted advertising, and news content tailoring.
	The challenges, however, lie in the limited amount of labeled data and the large scale of social networks. 
	In this paper, we formalize the problem of inferring location from social media into a semi-supervised factor graph model (SSFGM). 
    The model provides a probabilistic framework in which various sources of information (e.g., content and social network) can be combined together. 
    We  design a two-layer neural network to learn feature representations, and incorporate the learned latent features into SSFGM.
	To deal with the large-scale problem, we propose a Two-Chain Sampling (TCS) algorithm to learn  SSFGM. The algorithm achieves a  good trade-off between accuracy and efficiency. 
	Experiments on Twitter and Weibo show that the proposed TCS algorithm for SSFGM  can substantially improve the inference accuracy over several state-of-the-art methods. 
	More importantly, TCS achieves over $100\times$ speedup comparing with traditional propagation-based methods (e.g., loopy belief propagation).
	
	\hide{
	generalizes previous methods by incorporating content, network, and deep features learned from social context. The model is also flexible enough to support both supervised learning and semi-supervised learning.
	Second, we explore several learning algorithms for the proposed model, and present a Two-Chain Sampling (TCS) algorithm, which improves the inference accuracy.
	Third, we validate the proposed model on different genres of data -- Twitter and Weibo -- and demonstrate that the proposed model can substantially improve the inference accuracy (+3.3-18.5\% by F1-score) over that of several state-of-the-art methods.

	Third, we validate the proposed model on different genres of data -- Twitter and Weibo -- and demonstrate that the proposed model can substantially improve the inference accuracy 
over several state-of-the-art methods.
}
\end{abstract}



\maketitle

\section{Introduction}
\label{sec:intro}

In  social media platforms, such as Twitter, Facebook, and Weibo, location is a important demographic attribute to support friend and message recommendation~\cite{backstrom2010find,mok2007did}.
For example, statistics show that the average number of friends between users from the same time
zone is about 50 times higher than the number between users with
a distance of three time zones~\cite{Hopcroft:11CIKM}.
This geographical information, however, is usually unavailable. 
Cheng et al.~\cite{cheng2010you} show 
that only 26.0\% of users on Twitter input their locations. 
Furthermore, of the locations that are user-supplied, many are ambiguous or incorrect.
Twitter, Facebook, and Weibo 
have functionalities allowing per-tweet geo-tags; however, it turns out that only 0.42\% of all tweets contain a geo-tag~\cite{cheng2010you}.

In this work, we aim to find an \textit{effective} and \textit{efficient} way to automatically infer 
users' geographical locations from social media data. 
Different from previous works~\cite{backstrom2010find,cheng2010you,li2012towards} that deal with this problem in a specific scenario (e.g., determining the US cities only) or with specific data (e.g., Twitter), we propose a method that is general enough to apply to diverse scenarios.
This brings several new challenges:

\begin{itemize}
	\item \textbf{Limited labeled data.} Only a small portion of users have the  location information, and all the others are unlabeled.
 It is necessary to design a principled way to learn with both labeled and large unlabeled data.
 
 \item \textbf{Large-scale network.} Our problem has strong network correlation, but how to leverage the correlation, particularly in a large-scale network is challenging.
 
 \item \textbf{Model flexibility.} The proposed model should be flexible enough in oder to be easily generalized to other scenarios and to incorporate various information (e.g., content, structure and deep features).
\end{itemize}


\para{Previous work on location inference.}
The location inference problem has been  studied by researchers from different communities. Surveys  of location inference techniques on Twitter and related data challenges can  be  found in~\cite{ajao2015survey,jurgens2015geolocation,han2016twitter}.
Roughly speaking, existing literature can be divided into two categories. The first category of research focuses on studying content. For example, Cheng et al.~\cite{cheng2010you} and Han et al.~\cite{han2012geolocation} used a probabilistic framework and illustrated how to find local words and overcome tweet sparsity. 
Eisenstein et al.~\cite{eisenstein2010latent} proposed the Geographic Topic Model to predict a user's geo-location from text and topics. 
Ryoo et al.~\cite{ryoo2014inferring} applied a similar idea to a Korean Twitter dataset.
Ikawa et al.~\cite{ikawa2012location} used a rule-based approach to predict a user's current location based on former tweets.
Wing et al.~\cite{wing2011simple} and Roller et al.~\cite{roller2012supervised} proposed information retrieval approaches with geographic grids.  
The other line of research  infers user locations using network structure information. For example, Backstrom et al.~\cite{backstrom2010find} assumed that an unknown user would be co-located with one of their friends and sought the location with the maximum probability. 
McGee et al.~\cite{mcgee2013location} integrated social tie strengths between users to improve location estimation.
Jurgens~\cite{jurgens2013s} and Davis Jr et al.~\cite{zubiaga2016towards} used the idea of label propagation to infer user locations according to their network distances from users with known locations. These methods, however, do not consider content. Li et al.~\cite{li2012towards} proposed a unified discriminative influence model and utilized both the content and the social network, but they focused on the US users and only considered the mentioned location names in tweets. Rahimi et al. \cite{rahimi2015exploiting,rahimi2015twitter} used a simple hybrid approach to combine predictions from content and network. 
Recently, Miura et al. \cite{miura2017unifying} proposed a recurrent neural network model for learning content representations and integrated the user network embeddings.  
Another study using user profiles can be found in \cite{zubiaga2016towards}.
Table~\ref{related}  summarizes the most  related works on location inference. 
However, all the aforementioned methods cannot solve all the challenges listed above.

\begin{table}
\centering
\caption{\label{related}Summary of previous studies on geo-location inference in social media.}
\vspace{-0.05in}
\begin{tabular}{@{\;}c|c@{\;\;}|c@{\;\;}|c@{\;}|c@{\;}}
\toprule
\textbf{Features} & \textbf{Authors} & \textbf{Scope} & \textbf{Language} & \textbf{Object} \\\midrule
\multirow{6}{*}{Content} 
& Cheng et al.~\textsuperscript{\cite{cheng2010you}} & US & All & User \\
& Han et al.~\textsuperscript{\cite{han2012geolocation}} & World & English & User \\
& Eisenstein et al.~\textsuperscript{\cite{eisenstein2010latent}} & US & All & User \\
& Ryoo et al.~\textsuperscript{\cite{ryoo2014inferring}} & Korea & Korean & User \\
& Ikawa et al.~\textsuperscript{\cite{ikawa2012location}} & Japan & Japanese & Tweet \\
& Wing et al.~\textsuperscript{\cite{wing2011simple}} & US & English & User \\
& Roller et al.~\textsuperscript{\cite{roller2012supervised}} & US & English & User \\
\hline
\multirow{4}{*}{Network}
& Backstrom et al.~\textsuperscript{\cite{backstrom2010find}} & World & All & User \\
& McGee et al.~\textsuperscript{\cite{mcgee2013location}} & World & All & User \\
& Jurgens~\textsuperscript{\cite{jurgens2013s}} & World & All & User \\
& Davis Jr et al.~\textsuperscript{\cite{davis2011inferring}} & World & All & User \\\hline
\multirow{2}{0.5in}{Content + Network}
& Li et al.~\textsuperscript{\cite{li2012towards}} & US & English & User\\
& Miura et al.~\textsuperscript{\cite{miura2017unifying}} & World & English & User \\\hline
Profile
& Zubiaga et al.~\textsuperscript{\cite{zubiaga2016towards}} & World & All & Tweet \\\bottomrule
\end{tabular}
\end{table}

\vpara{Problem formulation.}
We now give a formalization to precisely define the problem we are dealing with.
Without loss of generality, our input can be considered as a \textit{partially labeled network} $G=(V, E, \mathbf{X}, Y^L)$ derived from the social media data. $V$ denotes a set of $|V|=N$ users, $V^L \subset V$ denotes a subset of labeled users (with locations), $V^U=V\setminus V^L$ indicates the subset of unlabeled users (without locations), $E \subseteq V\times V$ is the set of relationships between users, $Y^L$ corresponds to the locations of users in $V^L$, 
and $\textbf{X}$ is the feature matrix associated with users in $V$, where each row corresponds to a user and each column corresponds to a feature. 
Given the input, the problem of inferring user locations can be defined as follows:

\hide{
\begin{definition}{\textbf{\textit{Partially labeled network}}.}
	A \textit{partially labeled network} is an augmented social network denoted as $G=(V, E, Y^L, \mathbf{X})$, where $V = V^L \cup V^U$ denote the user set, $V^L$ is the set of labeled users and $V^U$ is the set of unlabeled users; $E \subseteq V\times V$ is the set of relationships between users; $Y^L$ is the set of labels (locations) corresponding to the users in $V^L$; $\textbf{X}$ is the feature matrix associated with users in $V$ where each row represents a user, each column represents a feature, and $x_{ij}$ denotes the value of $j^{th}$ feature of user $v_i$.
\end{definition}
}


\begin{problem}{\textbf{\textit{Geo-location Inference}}.} 
	Given a \textit{partially labeled network} $G=(V, E, \mathbf{X}, Y^L)$, the objective is to learn a predictive function $\mathcal{F}$ in order to predict the locations of unlabeled users $V^U$
	\begin{equation}
	\mathcal{F}: G=(V, E, \mathbf{X}, Y^L) \rightarrow Y^U
	\end{equation}
	\noindent where $Y^U$ is the set of predicted locations for unlabeled users $V^U$. 
\end{problem}


It is worth noting that our formulation of user location inference is slightly different from that in the aforementioned work.
The task is defined as a semi-supervised learning problem for networked data --- we have a network with limited labeled nodes and a large number of unlabeled nodes. 
Our goal is to leverage both local attributes $\mathbf{X}$ and network structure $E$ to
learn the predictive function $\mathcal{F}$.
Moreover, we assume that all predicted locations are among the locations occurring in the labeled set $Y^L$.
It is also worth mentioning that a user may have multiple locations; here we focus on predicting one's
primary location (e.g., home or the location  in one's profile). 

\vpara{Our solution and contributions.}
In this paper, we propose a probabilistic framework based on factor graphs to address the location inference problem. 
However, it is infeasible to directly apply traditional factor graphs, due to the new challenges in our problem.
Our goal is to achieve a good trade-off between the accuracy and efficiency, and also to make the model scalable to large networks.
Our contributions in this work can be summarized as follows:

\begin{itemize}
	\item  We present a semi-supervised factor graph model (SSFGM), which learns to infer user locations using both labeled and unlabeled data.
	
	\item By incorporating network structures and deep feature representations, SSFGM substantially improves the inference accuracy over several state-of-the-art methods. 
	
	\item  We propose a Two-Chain Sampling (TCS) algorithm to learn the SSFGM. TCS achieves over $100\times$ speedup comparing with the traditional  loopy belief propagation method. All codes and data used in  this work are publicly available.\footnote{\url{https://github.com/thomas0809/SSFGM}}
	
\end{itemize}

\hide{
The model seamlessly combines content information and network structure into a probabilistic graphical model. In the graphical model, labeled locations are propagated to unlabeled users. In this way, the model supports both supervised learning and semi-supervised learning. 
The model also supports incorporating deep representation features learned from social context.
To improve the learning effectiveness and efficiency, we  explore several learning algorithms. It shows that traditional Loopy Belief Propagation (LBP) algorithm is not scalable to large networks. The Softmax regression (SR) and Two-Chain Sampling (TCS) algorithms we proposed have successfully addressed the efficiency problem and improved the inference accuracy. 
}

We conduct systematic experiments on different genres of datasets, including Twitter and Weibo. 
The results show that the proposed  model significantly improves the location inference accuracy. In terms of the time cost for model training, the proposed TCS algorithm is very efficient and requires only less than two hours of training on million-scale networks. 

\section{Probabilistic Framework}
\label{sec:approach} 

In this section, we propose a semi-supervised  framework based on factor graphs for 
 location inference from social media. 

\subsection{Semi-supervised Factor Graph (SSFGM)}
\label{subsec:model}

\begin{figure*}[t]
\centering
\includegraphics[height=2.94in]{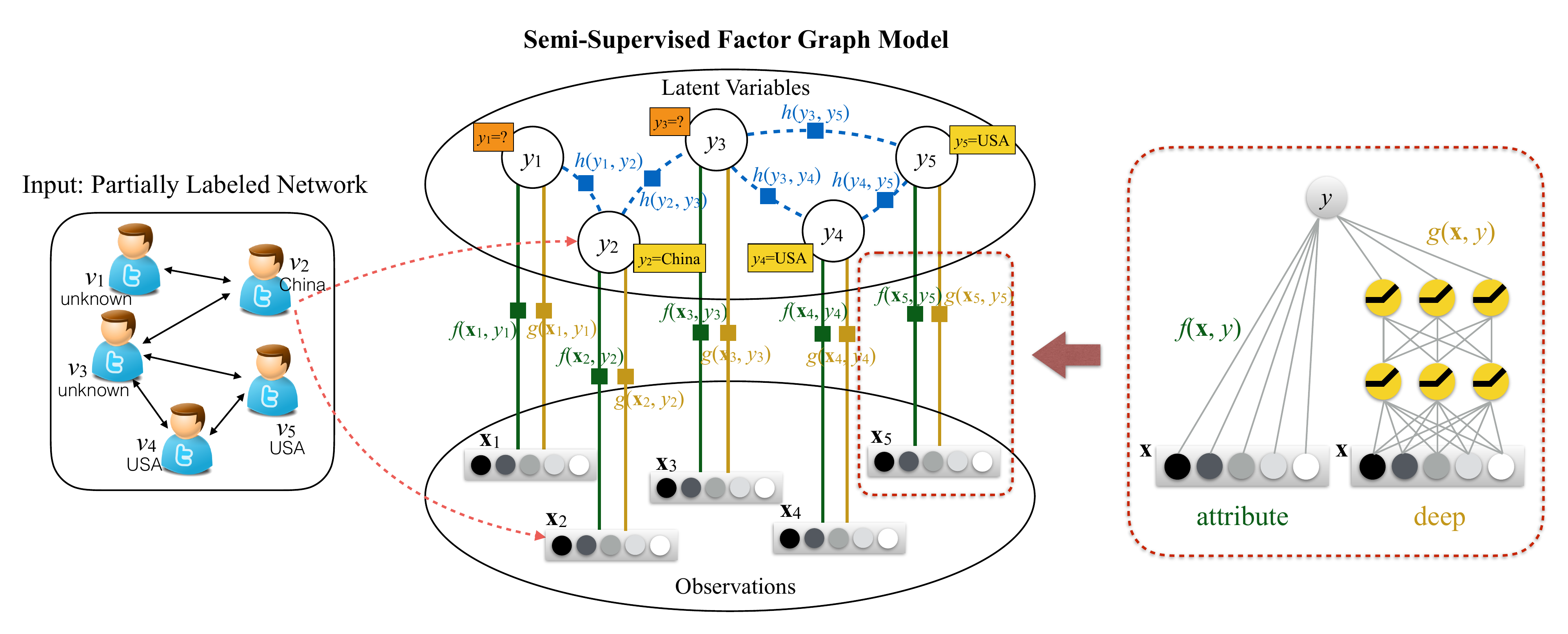}
\vspace{-0.0in}
\caption{\label{fig:fgm}Graphical representation of the proposed Semi-Supervised Factor Graph Model (SSFGM).}
\end{figure*}

\vpara{Basic intuitions.}
For inferring user locations, we have three basic intuitions. First, the user's profile may contain implicit information about the user's location, such as the time zone and the language selected by the user. Second, the tweets posted by a user may reveal the user's location. 
For example, Table \ref{table:content} lists the most popular ``local'' words in five English-speaking countries. These words include cities (Melbourne, Dublin), organizations (HealthSouth, UMass), sports (hockey, rugby), local idiom (Ctfu, wyd, lad),  etc. 
Third, network structure can be very helpful for geo-location inference. In Twitter, for example, users can follow each other, retweet each other's tweets, and mention other users in their tweets. 
The principle of \emph{homophily} \cite{Lazarsfeld:54}
--- ``birds of a feather flock together'' \cite{McPherson:01} --- suggests that 
these ``connected'' users 
may come from the same place. 
This tendency was observed between Twitter reciprocal 
friends in~\cite{Hopcroft:11CIKM,kwak2010twitter}. Moreover, we found that the homophily phenomenon also exists in the mention network.
Table~\ref{table:mention} shows the statistics for US, UK, and China Twitter users.
We can see that when user A mentions (@) user B, the probability that A and B come from the same country is significantly higher than that they come from different countries. 
Interestingly, when a US user A mentions another user B in Twitter, the chance that user B is also from the US is  95\% , while if user A comes from the UK, the probability sharply drops to 85\%, and further drops to 80\% for users from China.
We also did  statistics on the US users at state-level and found that there is an 82.13\% chance that users A and B come from the same state if one mentions the other in her/his tweets.


\begin{table}[t]
\centering
\small
\caption{\label{table:content}Popular location indicative words in tweets posted by users from different countries.}
\vspace{-0.05in}
\begin{tabular}{@{}m{0.38in} | m{0.5in} | m{0.4in} | m{0.45in} | m{0.45in} | m{0.35in}}
\toprule
\textbf{Country} & \textbf{US} & \textbf{UK} & \textbf{Canada} & \textbf{Australia} & \textbf{Ireland} \\\midrule
\multirow{10}{0.35in}{\textbf{Top-10 words}*}
&HealthSouth & Leeds    & Calgary      & Melbourne & Dublin \\
&UMass          & Used     & Toronto       & Sydney      & Ireland \\
&Montefiore    & Railway & Vancouver  & Australia    & Irish \\
&Ctfu              & xxxx       & Ontario       & 9am           & Hum \\
&ACCIDENT   & whilst     & Canadian   & Type          & lads \\
&Panera         & listed      & Canada      & \celsius      & lad \\
&MINOR         & Xx          & BC              & hPa           & xxx \\
&wyd               & Xxx        & hockey       & Centre        & rugby \\
&Kindred         & tbh         & Available    & ESE            & Xxx \\
&hmu              & xx          & NB              & mm              & xxxx \\
\bottomrule
\multicolumn{6}{l@{}}{\footnotesize* Top-10 by mutual information \cite{yang1997comparative}, among the words occurred $>5000$ times. }
\end{tabular}
\end{table}

\begin{table}[t]
\centering
\small
\setlength{\tabcolsep}{4pt}
\caption{\label{table:mention} Who will Twitter user @? {\small(User A mentions User B)}}
\vspace{-0.05in}
\begin{tabular}{@{\;}l | lr @{\;\;}| lr @{\;\;}| lr @{\;}}
\toprule
User A & \multicolumn{2}{c|}{US} & \multicolumn{2}{c|}{UK} & \multicolumn{2}{c}{China} \\\midrule  
\multirow{5}{*}{User B} 
             & US & 95.05 \%  & UK & 85.69 \%      & China & 80.37 \% \\
            & Indonesia & 0.77 \%   & US & 5.12 \%      & Indonesia & 7.89 \% \\
            & UK & 0.75 \%     & Nigeria & 3.03 \%  & US & 5.97 \%  \\
            & Canada & 0.61 \% & Indonesia & 1.00 \%  & Korea & 0.96 \% \\
            & Mexico & 0.27 \%  & Ireland & 0.49 \% & Japan & 0.71 \% \\
\bottomrule
\end{tabular}
\end{table}

\vpara{Model illustration.}
Based on the above intuitions, we propose a Semi-Supervised Factor Graph Model (SSFGM) for location inference. Figure~\ref{fig:fgm} shows the graphical representation of the SSFGM. The graphical model SSFGM consists of two kinds of variables: observations $\{\mathbf{x}\}$ and latent variables $\{y\}$. In our problem, each user $v_i$ corresponds to an observation $\mathbf{x}_i$ and is also associated with a latent variable $y_i$. The observation $\mathbf{x}_i$ represents the user's personal attributes and tweet content, and the latent variable $y_i$ represents the user's location. 
In this paper, we consider location inference as a classification problem, i.e., $y_i\in\{1,\dots, C\}$ which can be the  user's country, state, or city, and $C$ is the number of possible location categories.
We denote $Y = \{y_1, y_2, \dots, y_N\}$, and $Y$ can be divided into a labeled set $Y^L$ and an unlabeled set $Y^U$. 
The latent variables $\{y_i\}_{i=1,\cdots,N}$ are correlated with each other, representing the social relationships between users. In SSFGM, such correlations can be defined as factor functions.

Now we explain the SSFGM in detail. 
Given a partially labeled network as input,  we define  two factor functions:

\begin{itemize}
\item \textbf{Attribute factor:} $f(\mathbf{x}_i, y_i)$ represents the relationship between  observation (features) $\mathbf{x}_i$ and the latent variable $y_i$;

\item \textbf{Correlation factor:} $h(y_i, y_j)$ denotes the correlation between the locations of users $v_i$ and $v_j$. 
\end{itemize}

The factor functions can be instantiated in different ways. In this paper, we  define the attribute factor as an exponential-linear function
\begin{equation}
\begin{split}
& f(\mathbf{x}_i, y_i) = \exp\left( \alpha^\top \Phi(\mathbf{x}_i, y_i) \right) \\
& \Phi_{k} (\mathbf{x}_i, y_i) = \mathds{1}{(y_i=k)}~\mathbf{x}_{i}, \quad k\in\{1,\dots, C\}
\end{split}
\end{equation}

\noindent where $\alpha=(\alpha_{1}, \cdots, \alpha_{C})^\top$ is the weighting vector, $\Phi=(\Phi_{1}, \cdots, \Phi_{C})^\top$ is the vector of feature functions,
and $\mathds{1}{(y_i=k)}$ is an indicator function which is equal to 1 when $y_i=k$ and 0 otherwise.

The correlation factor is defined as
\begin{equation}
h(y_i, y_j) = \exp\left( \gamma^\top \Omega(y_i, y_j) \right)
\end{equation}

\noindent where $\gamma$ is also a weighting vector, and $\Omega$ represents feature functions $\Omega_{kl}(y_i, y_j)=\mathds{1}{(y_i=k, y_j=l)}~\mathbf{w}_{ij}$, $\mathbf{w}_{ij}$ can be any features associated with users $v_i$ and $v_j$, such as the number of interactions. Correlation can be  directed (e.g., mention), or undirected (e.g.,  reciprocal follow, Facebook friend). For undirected correlation, we need to guarantee $\gamma_{kl}=\gamma_{lk}$ in the model.

\vpara{Model enhancement with deep factors.}
We introduce how to utilize deep neural networks to enhance the proposed SSFGM. It also demonstrates the flexibility of the model. 
\hide{
	Recently, deep neural networks have achieved 
 excellent performance in various fields including image classification~\cite{Krizhevsky2012ImageNet} and phrase representations~\cite{Cho2014Learning}. 
The proposed SSFGM is flexible enough to incorporate deep neural networks. 
Inspired by the wide and deep learning~\cite{cheng2016wide} , w
}%
We incorporate a deep factor $g(\mathbf{x}_i, y_i)$ in  SSFGM 
to represent the deep (non-linear) association between $\mathbf{x}_i$ and $y_i$.  
The right side of Figure~\ref{fig:fgm} illustrates how we combine the predefined attributes and the deep factor in SSFGM.

Specifically, our deep factor is a two-layer neural network. The input vector $\mathbf{x}$ is fed into a neural network with two fully-connected layers, denoted $\mathbf{h}_1(\mathbf{x})$ and $\mathbf{h}_2(\mathbf{x})$:
\begin{equation}
\begin{split}
 \mathbf{h}_1(\mathbf{x})&=\text{ReLU}(\mathbf{W}_1\mathbf{x}+\mathbf{b}_1)\\
 \mathbf{h}_2(\mathbf{x})&=\text{ReLU}(\mathbf{W}_2\mathbf{h}_1(\mathbf{x})+\mathbf{b}_2)
\end{split}
\end{equation}

\noindent where $\mathbf{W}_1,\mathbf{W}_2,\mathbf{b}_1,\mathbf{b}_2$ are parameters of the neural network, and we use $\text{ReLU}(x)=\max(0,x)$ \cite{glorot2011deep} as the activation function. Similar to the definition of attribute factor, we define
\begin{equation}
\begin{split}
& g(\mathbf{x}_i, y_i) = \exp\left( \beta^\top \Psi(\mathbf{x}_i, y_i) \right) \\
& \Psi_{k} (\mathbf{x}_i, y_i) = \mathds{1}{(y_i=k)}~\mathbf{h}_2(\mathbf{x}_i), \quad k\in\{1,\dots, C\}\end{split}
\end{equation}

\noindent where $\beta$ is the weighting vector for the output of the neural network.

Thus, we define the following 
joint distribution over $Y$:
\begin{equation}
p(Y|\mathbf{X}) = \frac{1}{Z} \prod_{v_i \in V} f(\mathbf{x}_i, y_i) g(\mathbf{x}_i, y_i) \prod_{(v_i, v_j) \in E} h(y_i, y_j)
\label{eqn:joint}
\end{equation}
\noindent where $Z$ is the normalization factor that ensures $\sum_Y p(Y|\mathbf{X}) = 1$.

\vpara{Feature Definitions.}
For the attribute factor, 
we define two categories of features: profile and content. 

\textbf{Profile features} include information from the user profiles, such as time zone, user-selected language, gender, age, number of followers and followees, etc. 

\textbf{Content features} capture the characteristics of tweet content. The easiest way to define content features is using a bag-of-words representation. But it suffers from sparsity and high dimensionality, especially in Twitter, which has hundreds of languages. 

In our work, we employ Mutual Information (MI)~\cite{yang1997comparative} to represent the content. Given a word $w$ and a location $c$, the Mutual Information between them is computed as
\begin{equation} 
\text{MI}(w,c)=\log \frac{p(w,c)}{p(w)p(c)} \approx \log \frac{\text{count}(w,c)\cdot n}{\text{count}(w)\cdot\text{count}(c)}
\end{equation}

\noindent where $\text{count}(w,c)$ is the number of tweets which are posted at location $c$ and contain the word $w$, $\text{count}(w)$ is the number of tweets containing word $w$,  $\text{count}(c)$ is the number of tweets posted at location $c$, and $n$ is the total amount of tweets in the training data.
We pre-compute the MI between each word and each location using the training corpus, and define the content features for each user as the aggregated MI. We use two aggregation approaches, $max$ and $average$, i.e., 
\begin{equation}
\begin{split}
& \text{MI}_{max}(v,c) = \max_{w\in T(v)} \text{MI}(w,c) \\
& \text{MI}_{average}(v,c) = \frac{1}{|T(v)|} \sum_{w\in T(v)} \text{MI}(w,c)
\end{split}
\label{eqn:content}
\end{equation}

\noindent where $T(v)$ represents all the words from the tweets posted by user $v$. Then we use the aggregated MIs as the input content features for our model.



\subsection{Two-Chain Sampling (TCS) Learning} 
\label{sec:alg}
Now we introduce how to tackle the learning problem in SSFGM. We first start with the learning objective and gradient derivation,
and then propose our Two-Chain Sampling algorithm.

\vpara{Learning objective and gradient derivation.}
Learning a Semi-Supervised Factor Graph Model involves two parts: learning parameters $\alpha, \beta, \gamma$ for the graphical model, and learning parameters $\mathbf{W}, \mathbf{b}$ for the neural network of the deep factor. In this paper, we learn the two parts jointly. 

We follow the maximum likelihood estimation (MLE) to learn the graphical model. For notation simplicity, we rewrite the joint probability (Eq.~\ref{eqn:joint}) as 
\begin{equation}
\begin{split}
p(Y|\mathbf{X})&= \frac{1}{Z}\prod_i \exp\left(\theta^\top\mathbf{s}(y_i)\right) 
=\frac{1}{Z} \exp\left(\theta^\top\mathbf{S}(Y)\right)
\end{split}
\label{eqn:joint2}
\end{equation}

\noindent where  $\theta=(\alpha ^\top, \beta^\top, \gamma^\top)^\top$ are the factor graph model parameters to estimate, $\mathbf{s}(y_i) = (\Phi(\mathbf{x}_i, y_i)^\top, \Psi(\mathbf{x}_i, y_i)^\top, \sum_{y_j} \Omega(y_i, y_j)^\top)^\top$, and $\mathbf{S}(Y)=\sum_i \mathbf{s}(y_i)$. 
The input of SSFGM is partially labeled, which makes the model learning very challenging. The general idea here is to maximize the marginal likelihood of labeled data. We denote $Y|Y^L$ as the label configuration that satisfies all the known labels. Then we can define the following MLE objective function $\mathcal{O}(\theta)$:
\begin{equation}
\begin{split}
\mathcal{O}(\theta) & = \log p(Y^L|\mathbf{X}) = \log \sum_{Y|Y^L} \frac{1}{Z}\exp(\theta^\top\mathbf{S}) \\
& = \log \sum_{Y|Y^L} \exp(\theta^\top\mathbf{S}) - \log\sum_{Y} \exp(\theta^\top\mathbf{S})
\end{split}
\label{eq:mle}
\end{equation}

Now the learning problem is cast as finding the best parameter configuration that maximizes the objective function, i.e., 
\begin{equation}
\hat{\theta} = \arg \max_\theta \log p(Y^L|\mathbf{X})
\label{equ:learning}
\end{equation}

We can use gradient descent to solve this optimization problem. First, we derive the gradient of  parameter $\theta$:
\begin{equation}
\begin{split}
\frac{\partial \mathcal{O}(\theta)}{\partial \theta} 
&= \frac{\sum_{Y|Y^L}\exp (\theta^\top\mathbf{S}) \cdot \mathbf{S}}{\sum_{Y|Y^L}\exp (\theta^\top\mathbf{S})} 
- \frac{\sum_{Y}\exp (\theta^\top\mathbf{S}) \cdot \mathbf{S}}{\sum_{Y}\exp (\theta^\top\mathbf{S})} \\
&= \mathbb{E}_{p_\theta(Y|Y^L, \mathbf{X})} [\mathbf{S}] - \mathbb{E}_{p_\theta(Y|\mathbf{X})} [\mathbf{S}] \\
\end{split}
\label{eqn:gradient}
\end{equation}

In order to learn the neural network parameters in the deep factor, we derive the gradients of the top layer of the neural network similarly to Eq.~\ref{eqn:gradient}, and then follow the standard backpropagation algorithm to update the parameters. Similar methods have been studied in \cite{artieres2010neural}; we mainly discuss how to learn the graphical model in the following. 

In Eq.~\ref{eqn:gradient}, the gradient is equal to  the difference of two expectations under two different distributions. The first one --- $p_\theta(Y|Y^L, \mathbf{X})$ --- is the model distribution conditioned on labeled data, 
and the second --- $p_\theta(Y|\mathbf{X})$ --- is the unconditional model distribution. 
Both of them are intractable and cannot be computed directly \cite{sutton2006introduction}. 
We will illustrate how to deal with this challenge in the rest of the section.


\vpara{Loopy Belief Propagation (LBP) \cite{murphy1999loopy}.}
A traditional approach is LBP, an algorithm for approximately estimating marginal probabilities in graphical models. 
It performs message passing between variable nodes and factor nodes according to the sum-product rule \cite{kschischang2001factor}. In each step of gradient descent, we need to perform LBP twice to estimate $p_\theta(Y|\mathbf{X})$ and $p_\theta(Y|Y^L, \mathbf{X})$ respectively, and then calculate the gradient according to Eq.~\ref{eqn:gradient}. 

However, the LBP-based learning algorithm is computationally expensive. Its time complexity is $O(I_1I_2(|V|C+|E|C^2))$, where $I_1$ is the number of iterations for gradient descent, $I_2$ is the number of iterations for loopy belief propagation, and $C$ is the number of the location categories (usually 30-200). 
This algorithm is very time-consuming,
and not applicable especially when we have millions of users and edges.

\vpara{Softmax Regression (SR).} We try to solve the learning challenge in large-scale factor graphs.
It is difficult to calculate the joint probability Eq.~\ref{eqn:joint2}  because of the normalization factor $Z$, which sums over all the possible configurations of $Y$.
However, if we only consider a single variable $y_i$ and assume all the other variables are fixed, its conditional probability can be easily calculated by a softmax function,
\begin{equation}
p(y_i|\mathbf{X}, Y\setminus \{y_i\})=\frac{\exp\left(\theta^\top\mathbf{s}(y_i)\right) }{\sum_{y_i'}\exp\left(\theta^\top\mathbf{s}(y_i')\right)} \label{eqn:softmax}
\end{equation}

Eq.~\ref{eqn:softmax} has the same form as softmax regression (also called multinomial logistic regression). The difference is that the neighborhood information is captured in feature function $\mathbf{s}(y_i)$. Softmax regression can  be trained using gradient descent, and the gradient is much easier to compute than factor graph models. We then design an approximate learning algorithm based on softmax regression:
\begin{enumerate}[\quad Step 1.]
\item Conduct softmax regression to learn $\alpha$ and $\beta$, with labeled data $\{(\mathbf{x}_i, y_i)|y_i\in Y^L\}$ only;\footnote{Here we assume $p(y_i|\mathbf{x}_i)=\text{softmax}\left(\alpha^\top \Phi(\mathbf{x}_i, y_i) + \beta^\top\Psi(\mathbf{x}_i, y_i) \right)$. }
\item Predict the labels $Y^U$ for unlabeled users;
\item Conduct softmax regression to learn $\theta$ according to Eq.~\ref{eqn:softmax};
\item Predict the labels  $Y^U$ for unlabeled users. If the prediction accuracy on the validation set increases, go to Step 3; otherwise, stop.
\end{enumerate}

This algorithm is an efficient approximation method for learning SSFGM, but its performance can be further improved. We can use SR to initialize the model parameters for the other learning algorithms.

\begin{algorithm}[t]
	\caption{\label{TCS}Two-Chain Sampling (TCS)}
	\SetKwInOut{Input}{Input}
	\SetKwInOut{Output}{Output}
	\SetKwRepeat{Repeat}{repeat}{until}
	\Input{ $G=(V, E), \mathbf{X}, Y^L$, learning rate $\eta$;} 
	\Output{ learned parameters $\theta$;}
	Initialize $\theta$ randomly\; 
	Initialize $Y_1$ with $Y^L$ fixed, and $Y^U$ randomly\;
	Initialize $Y_2$ randomly\;
	\Repeat{early stopping criteria satisfied}{
		Randomly split $V$ to mini-batches $\{B_1,\dots,B_K\}$\;
		\For {$k=1,2,\dots, K$} {
			Initialize the gradient $\delta \leftarrow \mathbf{0}$\;
			\For {$v_i\in B_k$} {
				Sample $y_i$ in $Y_1$ such that $Y_1\sim p_\theta(Y|\mathbf{X},Y^L)$\;
				Sample $y_i$ in $Y_2$ such that $Y_2\sim p_\theta(Y|\mathbf{X})$\;
				\If{ $y_i \in Y^L$ }
				{$\delta \leftarrow \delta + \mathbf{s}(y_i|Y_1) 
				- \mathbb{E}[\mathbf{s}(y_i) | Y_2\backslash\{y_i\}]$\;}
				\Else
				{$\delta \leftarrow \delta + \mathbb{E}[\mathbf{s}(y_i) | Y_1\backslash\{y_i\}] 
				- \mathbb{E}[\mathbf{s}(y_i) | Y_2\backslash\{y_i\}]$\;}
			}
			$\theta\leftarrow \theta + \eta\cdot \delta$\;
		}
		Evaluate on the validation set\;
	}
\end{algorithm}

\vpara{Two-Chain Sampling (TCS).} 
Now we introduce the proposed TCS algorithm,  a novel 
Markov Chain Monte Carlo (MCMC) method \cite{Andrieu:03}, for efficiently learning SSFGM. 
MCMC has been proven successful in learning complex graphical models. For example, Rohanimanesh et al. proposed the SampleRank algorithm to train factor graphs \cite{rohanimanesh2011samplerank}. 
However, SampleRank has some shortcomings. It actually optimizes an alternative max-margin objective instead of the original maximum likelihood objective.
In addition, it relies on an external metric (e.g., accuracy), which could be arbitrary and engineering-oriented, since multiple metrics are often available for evaluation. 

We propose a new method to directly optimize the maximum likelihood objective (Eq.~\ref{eq:mle}) without using  additional heuristic metrics. We refer to this algorithm as \textit{Two-Chain Sampling},  summarized in Algorithm~\ref{TCS}.  
The key idea behind TCS is that we generate two Markov chains, and in each sampling step, we use a similar approach as that of contrastive divergence (CD)~\cite{hinton2002training} to compute the gradient.

Mathematically, the gradient we are estimating (Eq.~\ref{eqn:gradient}) consists of two expectation terms. 
To obtain an unbiased estimation, we construct two Markov chains $Y_1$ and $Y_2$. Specifically, we sample $Y_1$ from $p_{\textit{data}} = p_\theta(Y|\mathbf{X}, Y^L)$ and sample $Y_2$ from $p_{\textit{model}} = p_\theta(Y|\mathbf{X})$. Various samplers could be applied here. We choose Gibbs sampling~\cite{geman1987stochastic} in this work.
\footnote{We also tried some other sampling methods such as Metropolis-Hastings sampling \cite{hastings1970monte}, and finally chose Gibbs sampling because of its efficiency.
}
In each sampling step, Gibbs sampling updates a single variable $y_i$ while the other variables are fixed. In other words, we sample $y_i$ according to the distribution we have defined in Eq.~\ref{eqn:softmax}, but use the neighbours' values from $Y_1$ and $Y_2$ respectively in the two chains. It should also be noted that when we update $y_i$ of a labeled user in the chain $Y_1$ (i.e., $y_i\in Y^L$), its value should never be changed from its true label. Since $Y_1$ follows $p_\theta(Y|\mathbf{X}, Y^L)$, all the known labels must be fixed.

It is non-trivial to calculate the gradient in the sampling process. A standard way is to keep sampling for a number of iterations and then use the resulting distribution to approximately compute the expectation value. However, the MCMC method typically requires too many iterations to reach convergence, which makes it not applicable in training large factor graph models. Fortunately, as suggested by the contrastive divergence algorithm~\cite{hinton2002training}, we do not have to wait for the convergence but usually a few sampling steps (or even one step) can be effective enough. 
Besides, bearing a similar merit to stochastic gradient descent (SGD)~\cite{bottou2010large}, we can sample only a small subset of variables each time instead of all of them. 
Thus we first randomly split the user set into some fix-sized mini-batches. In each step, we sample the variables $y_i$ in a mini-batch, compute the gradient, and update the parameters. The gradient can be approximated as $\sum_i \mathbf{s}(y_i|Y_1) - \mathbf{s}(y_i|Y_2)$, where the summation is taken over the mini-batch. Empirically, it is a feasible solution, but the learning process sometimes becomes unstable. To improve learning stability, we change the gradient computation to $\sum_i \mathbb{E}[\mathbf{s}(y_i) | Y_1\backslash\{y_i\}] - \mathbb{E}[\mathbf{s}(y_i) | Y_2\backslash\{y_i\}]$, i.e., the expectations under the distribution Eq.~\ref{eqn:softmax}. Again, the first expectation value is simply $s(y_i|Y_1)$ if $y_i$ is a known label. We have explicitly indicated it in Algorithm~\ref{TCS} with the ``if-then-else'' statement. In practice, it is usually necessary to downsample the unlabeled data if they are significantly more than the labeled data.

We use the early stopping technique to determine when to stop training. Specifically, we divide the labeled data into a training set and a validation set. During the learning process, we only use the labels in the training set. We evaluate the model after each epoch (a complete pass through the dataset), and if the prediction accuracy on the validation set does not increase for $\varepsilon$ epochs, we stop the algorithm and return the parameter configuration $\hat{\theta}$ that achieves the best accuracy on the validation set. $\varepsilon$ is a hyperparameter. 

Compared with LBP and SR, the TCS algorithm directly optimizes the MLE objective, and is very time-efficient. Focusing on the semi-supervised learning setting on a partially labeled factor graph, we simultaneously maintain two Markov chains and provide an elegant way to perform gradient estimation.

\vpara{Parallel learning.}
To scale up the proposed model to handle large networks, we have developed 
parallel learning algorithms for SSFGM. 
For the SR algorithm, softmax regression can be easily parallelized. The gradient is a summation over all the training instances (or a mini-batch if using SGD), and the computation is independent.
For TCS, we can still parallelize the computation of the instances in a mini-batch. The only difference is that instead of sampling the variables one by one in the sequential setting, we sample a mini-batch of variables simultaneously in the parallel setting. This variation is usually called the blocked Gibbs sampler~\cite{ishwaran2001gibbs} and will not change the original properties of Gibbs sampling.

\vpara{Prediction.}
SSFGM is learned in a semi-supervised way --- both labeled and unlabeled instances are taken as input in the training process. After learning the parameters, we predict the labels of unlabeled instances. Alternatively, 
we can also apply the learned SSFGM in a inductive setting, i.e., to predict future unknown instances.

For prediction, the task is to find the most likely configuration of $\hat{Y}$ for unlabeled users based on the learned parameters $\hat{\theta}$,
\begin{equation} 
\hat{Y} =\arg\max_{Y|Y^L} p_{\hat{\theta}}(Y|\mathbf{X}, Y^L)
\end{equation}

We also use the sampling method to obtain the predictions. In principle, we can keep sampling with the estimated $\hat{\theta}$ and return the configuration $\hat{Y}$ with the maximum likelihood. But in practice, 
we simply 
choose the value with the highest probability in each sampling step. 
It only guarantees finding a local optimum, but is usually effective enough and much faster. (Cf. \S~\ref{sec:exp} for details.)

\section{Experiments}
\label{sec:exp} 

We evaluate the proposed model on two different social media data:
Twitter and Weibo. 
 
 
 \begin{table}
 	\centering
 	\caption{\label{statistics}Statistics of the datasets.}
 	\begin{tabular}{p{0.9in}|r|r|r}
 		\toprule
 		\multicolumn{1}{c|}{\textbf{Dataset}} & \multicolumn{1}{c|}{\textbf{\#user}} & \multicolumn{1}{c|}{\textbf{\#edge}} & \multicolumn{1}{c}{\textbf{\#location}} \\ \midrule
 		Twitter (World) & 1,480,360 & 25,867,610  &  159 {\small $^{(a)}$} \\\hline
 		Twitter (USA) & 329,457 & 3,194,305 & 51 {\small $^{(b)}$} \\\hline
 		Weibo & 1,073,923 & 26,849,122 & 34 {\small $^{(c)}$} \\ 
 		\bottomrule
 		\multicolumn{4}{m{2.75in}}{\footnotesize $^{\star} $(a) 159 countries; (b) 50 states and Washington, D.C.; (c) 34 provinces.} 
 	\end{tabular}
 \end{table}

\subsection{Experimental Setup}

\vpara{Datasets.}
We construct three datasets for experiments.  Table \ref{statistics} shows the basic statistics of the datasets.

\begin{itemize}
\item \textbf{Twitter (World):} 
We collect  geo-tagged tweets posted in 2011 through Twitter API. There are 243,000,000 tweets posted by 3,960,000 users in our collected data. 
After data preprocessing, we obtain a dataset consisting of 1.5 million users from 159 countries in the world. The task on this dataset is to infer the user's country. 
Due to the limitations of the Twitter API, we cannot crawl the following relationships; thus we use  mentions (``@'') in tweets to derive the  relationships.

\item \textbf{Twitter (USA):} This dataset is constructed from the same raw data as that of Twitter (World). The difference is that we only keep the USA users here. The task on this dataset is to infer the user's state. 
\item \textbf{Weibo~\cite{zhang2013social}:} Weibo is the most popular Chinese microblog. 
The original dataset consists of about 1,700,000 users, with up to 1,000 of the most recent microblogs posted by each user. 
The task is to infer the user's province. We use reciprocal following relationships as edges in this dataset.

\end{itemize}


\begin{table*}[t]
	\centering
	\caption{\label{table:res} Performance comparison of different methods in user geo-location inference. (``Acc.'' means Accuracy (\%), and ``MED'' means Mean Error Distance (km).)}
	\vspace{-0.1in}
	\setlength{\tabcolsep}{6pt}
	\begin{tabular}{l|*{3}{c}|*{3}{c}|*{3}{c}}
	\toprule
	& \multicolumn{3}{c|}{Twitter (World)} & \multicolumn{3}{c|}{Twitter (USA)} & \multicolumn{3}{c}{Weibo} 
	\\\cmidrule{2-10}
	\multicolumn{1}{c|}{Method} & Acc. & Acc.@3 & MED &   Acc. & Acc.@3 & MED &   Acc. & Acc.@3 & MED 
	\\\midrule
		Content \textsuperscript{\cite{cheng2010you}} 
		&79.68 &91.01 &1278.89 
		&40.61 & 51.60 & 931.32
		&30.96 & 52.88 &  555.68 
		\\
		Logistic Regression
		& 94.44 & 98.18 & 302.06 
		& 48.22 & 67.37 & 707.34
		& 36.98 & 58.12 & 499.67 
		\\
		SVM \textsuperscript{\cite{zubiaga2016towards}}
		& 94.46 & 98.12 & 300.42 
		& 47.89 & 67.44 & 713.44 
		& 35.85 & 57.41 & 507.65  
		\\
		\midrule
		FindMe \textsuperscript{\cite{backstrom2010find}}
		& 83.46 & 86.99 & 1350.08  
		& 46.34 & 57.60 & 1314.85 
		& 63.92 & 81.00 & 281.14   
		\\ 
		GCN \textsuperscript{\cite{kipf2016semi}}
		& 94.54 & 97.98 & 288.20
		& 58.36 & 74.56 & 516.51 
		& 66.18 & 79.14 & 257.95
		\\
		\midrule
		SSFGM (SR) 
		& 95.18 & 98.29 & 280.87  
		& 56.12 & 73.15 & 606.28  
		& 64.32 & 80.27 & 281.61  
		\\
		SSFGM (SampleRank \textsuperscript{\cite{rohanimanesh2011samplerank}}) 
		& 94.96 & 98.25 & 292.15 
		& 58.48 & 74.54 & 578.95  
		& 66.91 & 82.81 & 263.29
		\\ 
		SSFGM (TCS) 
		& 95.68 & \textbf{98.32} & \textbf{229.77} 
		& 62.51 & 76.37 & 489.75 
		& \textbf{70.34} & 80.44 & 232.59  
		\\ 
		SSFGM (TCS+Deep)
		& \textbf{95.72} &{98.31}& 231.23 
		& \textbf{62.63} & \textbf{76.55} & \textbf{487.92} 
		& 70.06 & \textbf{82.89} & \textbf{231.39} 
		\\
		\bottomrule
	\end{tabular}
\end{table*}

We preprocess the three datasets in the following ways. First, we filter out users who have fewer than 10 tweets in the dataset. Then, we tokenize the tweet content into words. In Twitter, we split the sentences by punctuation and spaces. For languages that do not use spaces to separate words (such as Chinese and Japanese), we split each character. In the Weibo data provided by~\cite{zhang2013social},  the content has already been tokenized into Chinese words. For each user, we combine all her/his tweets and derive content features as defined in Eq.~\ref{eqn:content}. The ground truth location is defined by different ways in each dataset. In the two Twitter datasets, we convert the GPS-tag on tweets to its country/state, and only keep the users who posted all tweets in the same country/state in order to reduce the noise in the training data. (In our data, more than $90\%$ users posted all their tweets in the same country in a year, and more than $80\%$ USA users posted all their tweets in the same state.) In Weibo, the ground truth locations are extracted from user profiles, which have been categorized into provinces. We collect the latitude and longitude coordinates of the locations (for calculating the error distances) through the Google Maps Geocoding API.
In all datasets, we remove the countries/states/provinces with fewer than 10 users.

\vpara{Comparison methods.}
We compare the following methods for location inference:

\begin{itemize}
\item \textbf{Content \cite{cheng2010you}:} It utilizes a simple probabilistic model to predict locations with tweet content only.
\item \textbf{Logistic Regression (LR):} A baseline classification model to predict the user location using logistic regression. 
We use the same feature set as our proposed model, including both content and profile features, but ignoring the correlations. 

\item \textbf{Support Vector Machine (SVM)~\cite{zubiaga2016towards}:}  Zubiaga et al. have applied SVM to classify tweet location. We choose a linear function as the kernel of SVM.


\item \textbf{FindMe~\cite{backstrom2010find}:} This method infers user locations with social and spatial proximity. It uses the network only and propagates  label information to unlabeled users.

\item \textbf{Graph Convolutional Network (GCN)~\cite{kipf2016semi}:} We also consider GCN, a state-of-the-art neural network model for graph-based semi-supervised learning. It uses the same features and correlations as our model to predict user locations.

\item \textbf{SSFGM:} The proposed method. 
We compare the performance of our model trained by three different algorithms:  Softmax Regression (SR), SampleRank \cite{rohanimanesh2011samplerank}, and Two-Chain Sampling (TCS). 
We also report results when we enhance the model with deep factors: SSFGM~(TCS+Deep).
\end{itemize}

\para{Evaluation metrics.}
For evaluation, we divide each dataset into three parts: 50\% for training, 10\% for validation, and 40\% for testing. For the methods that do not require validation, the validation data is also used for training.
We consider three evaluation metrics: Accuracy (percentage of the users whose locations are predicted correctly), Accuracy@3 (percentage that the true location is among the top 3 predictions\footnote{All of the comparison methods can output a likelihood score for each location. We rank the locations according to the likelihood and evaluate the top 3.}), and Mean Error Distance (the average error distance between the prediction and the true location).

\vpara{Implementation details.}
For the Content method, we identify location indicative words using the Information Gain Ratio criterion proposed by \cite{han2012geolocation}.
For LR and SVM, we use the implementation of Liblinear~\cite{fan2008liblinear} with the default parameter setting. 
For GCN, we use a two-layer GCN model with the hidden layer size of 128, and use the mini-batched training approach \cite{chen2018fastgcn}.

For the proposed method, we implement 
SSFGM (TCS) and SSFGM (TCS+deep) using TensorFlow~\cite{abadi2016tensorflow} with the Adam optimizer~\cite{kingma2014adam}.
We empirically set up the hyperparameters according to the performance on the validation set. 
Specifically, we use a learning rate of $\eta=0.01$, a mini-batch size of 512, and an early stopping threshold of $\varepsilon=10$.
The deep factor is defined as a two-layer fully-connected neural network, where the first layer has 128 hidden units and the second layer has 64 hidden units.

All experiments are performed on an x86-64 machine with 40-core 3.00GHz Intel Xeon(R) CPUs, 3 NVIDIA  Titan X GPUs, and 128GB RAM.

\subsection{Experiment Results}

\vpara{Location inference performance.}
We compare the performance of all the methods on the three datasets. Table~\ref{table:res} lists the performance of comparison methods for geo-location inference. 

In our experiments, the proposed SSFGM consistently outperforms all the comparison methods in terms of prediction accuracy on all datasets. In Twitter (World), LR and SVM can achieve an accuracy of 94.4\% in predicting the user's country. 
Our SSFGM further improves the accuracy to 95.7\% by incorporating social network. 
In Twitter (USA) and Weibo, it becomes harder to predict a user's state/province. This is because for predicting user's country, the content information might already be very indicative, as users from different countries use different languages; while for predicting the state-level location, we need to exploit more information such as the social network. 
SSFGM achieves a significant improvement in comparison with other methods that only utilize local attributes or only utilize the network. It is noticeable that while using the same content  and network information, SSFGM significantly outperforms Graph Convolutional Network (GCN), 
a state-of-the-art method has been successfully applied in many 
other tasks on graphs. SSFGM directly models the correlation between the locations of related users, while GCN only models the correlation between the features. 
In fact, we have also tried to combine GCN and SSFGM by defining the attribute factor function using GCN (i.e., it takes the feature matrix $\mathbf{X}$ as input instead of a single user's feature $\mathbf{x}_i$ alone). However, it still cannot outperform SSFGM.


Another interesting discovery is that, in Twitter (USA), purely network-based methods (e.g., FindMe) perform worse than linear models (LR and SVM), but in Weibo (a Chinese microblog), they significantly outperform linear models. This suggests that network information is more important in the Weibo dataset. We suspect the reason might be the differences of user behaviours and population distributions between the USA and China.


Finally, we can observe that in general the deep factor helps to improve inference accuracy of our model. 
Our motivation to incorporate deep factor in our model is trying to capture the non-linear, high-dimensional association between input features and output locations. Although its benefit is not very significant in our experiments, we have shown the feasibility of using neural networks in our model. Designing more advanced and effective neural network architectures will be an interesting future direction.

\begin{table}[t]
\centering
\caption{\label{table:fb} Performance and training time of different learning algorithms for SSFGM on a small dataset \cite{snapnets}. (The numbers in brackets represent the speedup against LBP.)}
\vspace{-0.1in}
\begin{tabular}{l|c|c}
\toprule
Method & \multicolumn{1}{c|}{Accuracy} &  Time \\\midrule
LBP \textsuperscript{\cite{murphy1999loopy}}  & 91.52\% & 16.8 min   \\ 
SR & 90.94\% &  1.90 sec (530$\times$)  \\
SampleRank \textsuperscript{\cite{rohanimanesh2011samplerank}} & 91.23\% &  5.26 sec (192$\times$)  \\ 
TCS & 91.23\% & 8.57 sec (118$\times$)   \\ 
\bottomrule
\end{tabular}
\end{table}

\begin{table}[t]
\centering
\caption{\label{table:time} Training time of GCN and SSFGM.}
\vspace{-0.1in}
\begin{tabular}{@{}l@{\;}|@{\;}r@{\;}|@{\;}r@{\;}|@{\;}r@{}}
\toprule
Method & Twitter (World) & Twitter (USA) & \multicolumn{1}{c}{Weibo} \\\midrule
GCN & 11 hr 11 min &  48.3 min & 4 hr 18 min  \\
SSFGM (TCS) & 1 hr 55 min & 24.2 min & 47.6 min  \\
SSFGM (TCS+Deep) & 1 hr 57 min & 24.3 min & 1 hr 4 min \\
\bottomrule
\end{tabular}
\end{table}


\vpara{Comparison of different learning algorithms.}
Now we compare the performance of four different learning algorithms for SSFGM, including the traditional Loopy Belief Propagation (LBP) algorithm \cite{murphy1999loopy}.  LBP suffers from its high computational cost, and is not useful in our million-scale datasets. 
However, in order to fairly compare it with the other algorithms, we construct a smaller dataset with the Facebook ego-network data from SNAP~\cite{snapnets}. In this dataset, each user has an anonymized hometown location, but content information is not available. We use Facebook friendships as edges. After data preprocessing, we get a relatively small dataset with 856 users and 11,789 edges. 
Then we compare the performance of four learning algorithms on this dataset. Table~\ref{table:fb} shows the results, where the algorithms are mainly implemented in C++ and each one uses a single CPU core. Among the algorithms, LBP achieves the highest accuracy, but takes much more time to train than the others. The other three algorithms have significantly reduced the training time, either with approximation assumptions or sampling methods. SR seems to be the most time-efficient, but its accuracy is worse than that of the others. SampleRank and the proposed TCS algorithm solve the computation cost problem (over 100$\times$ speedup compared with LBP), and achieve comparable accuracy. From Table~\ref{table:res}, we can also see that TCS usually performs better than SampleRank on large datasets.

We report the training time of TCS on the three large datasets and compare them with GCN in Table~\ref{table:time}. Here the algorithms are running on three GPUs under the Tensorflow framework. With TCS, our model takes only 0.4--2 hours of training on million-scale datasets and achieves the best prediction performance among the comparison methods. It is also much faster than GCN.

\begin{figure}
\centering
\includegraphics[height=1.55in]{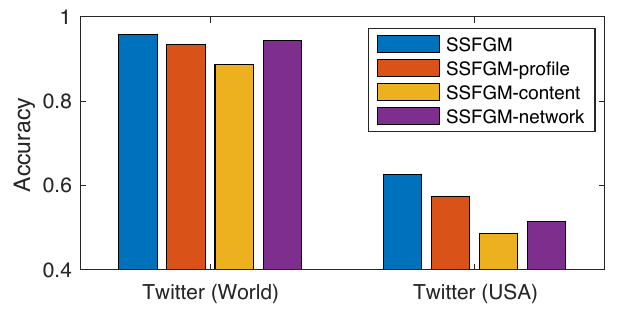}
\vspace{-0.1in}
\caption{\label{contrib}Feature contribution analysis. { (\text{SSFGM-profile}, \text{-content}, \text{-network} means removing profile features, content features, or correlation factors, respectively.)}}
\end{figure}

\begin{figure*}
\centering
\includegraphics[height=1.8in]{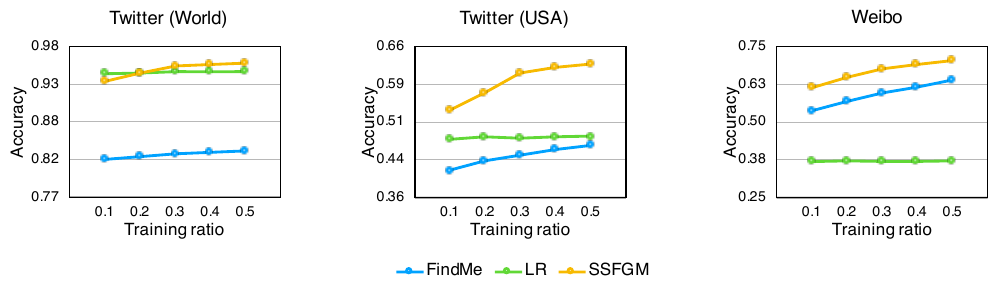}
\vspace{-0.05in}
\caption{\label{ratio}Training data ratio analysis.}
\end{figure*}

\vpara{Factor contribution analysis.} 
We evaluate 
how different factors (content, profile, and network) contribute to location inference in the proposed model. 
We use the two Twitter datasets in this study.
Specifically, we remove each factor from our SSFGM and then evaluate the model's prediction accuracy decrease. The larger the decrease, the more important the factor to the model.
\figref{contrib} shows the results on the Twitter datasets.
 We see that different  factors contribute
differently on the two datasets. The content-based features seem to be the most useful in the proposed model for inferring location on the Twitter datasets. On the other hand, all features are helpful. This analysis confirms the necessity of  incorporating various features
in the proposed model.

\vpara{Training data ratio analysis.} We conduct further experiments to evaluate our method's performance when training data is limited. We change the training data ratio in each dataset and compare several methods' prediction accuracies. The validation and testing sets remain constant. The results are shown in Figure~\ref{ratio}. SSFGM does quite well, even with only 10\% of labeled data. Its prediction accuracy steadily increases when more labeled data are used for training. It shows distinct advantages compared with LR, whose performance can hardly be improved by adding more training data.

\section{Conclusions}
\label{sec:conclusion} 

In this paper, we studied the problem of inferring user locations from social media. We proposed a general probabilistic  model based on factor graphs. The model generalizes previous methods by incorporating content, network, and deep features learned from social context. It is also sufficiently flexible to support semi-supervised learning with limited labeled data.
We proposed a Two-Chain Sampling (TCS) algorithm, which significantly improves the inference accuracy. This algorithm is also parallelizable  and is capable of handling large-scale networked data. 
Our experiments on three different datasets validated the effectiveness and the efficiency of the model. 

\hide{
Inferring user demographics from social media is a fundamental issue and represents a new
and interesting research direction. As for future work, it would
be intriguing to apply the proposed model to infer other demographic attributes such as gender and age. It is also interesting to connect the study to some real applications -- for example, advertising and recommendation -- to further evaluate how  inferred location can help real applications. 
Finally, we also consider how to integrate social theories such as social status and structural holes to understand the underlying mechanism behind user behavior and network dynamics.
}

\bibliographystyle{abbrv}
\bibliography{references}  

\end{document}